\documentclass[conference,letter]{IEEEtran}
\usepackage[utf8]{inputenc}
\usepackage[T1]{fontenc}

\usepackage{ifpdf}

\ifCLASSINFOpdf
  \usepackage[pdftex]{graphicx}
\else
\fi
\usepackage[cmex10]{amsmath}
\usepackage{amsfonts}

\usepackage{wrapfig}
\usepackage{url}
\usepackage{hyperref}

\usepackage{color}
\usepackage[table,usenames,dvipsnames]{xcolor}
\usepackage{subcaption}
\usepackage{bm}

\usepackage[placement=top,angle=0,color=black,scale=1,hshift=0,vshift=-10]{background}
\backgroundsetup{contents={Paper accepted for the ICDL-Epirob 2017 conference.}}

\hypersetup{
  colorlinks = false,
  pdfborder = {0,0,0},
  pdfauthor={Oswald Berthold and Verena V. Hafner},
  pdftitle={Tapping the sensorimotor trajectory},
  pdfsubject={Prediction learning},
  pdfkeywords={sensorimotor learning, reinforcement learning, machine learning},
  pdfproducer={pdflatex},
}

\begin{document}
%
\title{Tapping the sensorimotor trajectory}

\author{\IEEEauthorblockN{Oswald Berthold, Verena V. Hafner}
  \IEEEauthorblockA{Adaptive Systems Group, 
    Dept. of Computer Science, 
    Humboldt-Universität zu Berlin, Germany}
}

\IEEEoverridecommandlockouts


%


\newcommand{\FIXMET}[1]{\textbf{{\color{orange}FIXME}:} {\color{red}#1}}

\newcounter{para}

\newcommand{\parnum}{\textbf{\arabic{para}}}

\newcommand\myparamarh[1]{%
    \stepcounter{para}%
    \leavevmode\marginpar[\scriptsize\parnum\hspace{0.2pt} #1]{\scriptsize\parnum\hspace{0.2pt} #1}%
}

\newcommand\myparb[1]{}
\maketitle

\begin{abstract}
In this paper, we propose the concept of sensorimotor tappings, a new graphical technique that
explicitly represents relations between the time steps of an agent's
sensorimotor loop and a single training step of an adaptive internal model. 
In the simplest case this is a
\emph{relation linking two time steps}. In realistic cases these
relations can extend over several time steps and over different sensory
channels. The aim is to capture the footprint of information intake
relative to the agent's current time step. We argue that this view
allows us to make prior considerations explicit and then use them in
implementations without modification once they are established.

Here we explain the basic idea,
provide example tappings for standard configurations used in
developmental models, and show how tappings can be applied to problems
in related fields.

\end{abstract}


%
\IEEEpeerreviewmaketitle

\section{Introduction}
\label{sec:orge25b6d6}

\myparb{Learning as modelling} As stated in theories of development,
an agent's brain can contain modules that function as models of
its interaction with the world. These
models are used by the brain to evaluate the possible actions in
``imagined space'' and the agent only performs the most promising ones in
physical space. The role of a theory on these models is to describe how precisely
a sensorimotor model is learnt from experience and how it interacts
with other existing models in a developmental context.

\myparb{Levels of modelling} There are different types of such models. 
Machine learning (ML), for example, solves the
problem of fitting a model to data in a problem independent
form. The ML approach usually relies on a \emph{preprocessing step} to
transform the raw data into the required form. Using ML methods we can
learn sensorimotor models of transitions in sensorimotor space up to a 
desired accuracy. This level of modelling provides the grounding in sensorimotor
space. 
An important question is \emph{how to map the raw sensorimotor data to
  sensorimotor training data} for realizing specific functions needed
inside a developmental model.

\begin{figure}
  \centering
  \includegraphics[width=0.4\textwidth]{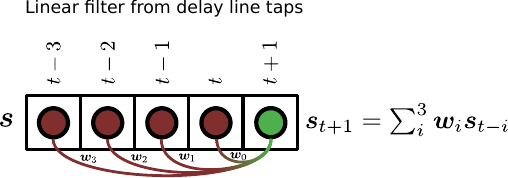}
    \caption{This graphical representation of linear a filter uses
      successively delayed copies of an input $\bm{s}$ to compute a
      prediction as a weighted sum of all copies. It provides the starting
    point for sensorimotor tappings.}
    \label{fig:ex-1-filter}
\end{figure}

This paper introduces the concept of \emph{tapping} for designing and
analysing models of developmental learning.  \myparb{Tappings simple}
The concept is adopted from signal processing where it is used to
describe a filter as a weighted sum of delayed copies of a signal as
shown in \autoref{fig:ex-1-filter}.  The simplest sensorimotor tapping
then is just the same as a filter tapping, using past values of a
single variable to predict a future value of the same variable. In
realistic situations the number of past values can be numerous,
include different modalities, and the linear filter is a general
nonlinear function whose parameters are learned from data. This view
allows us to discuss a wide range of issues in temporal learning. For
example, concepts from developmental robotics, reinforcement learning,
neuroscience, and information theory can be represented and compared
by exposing relational properties independent of terminology.

\begin{figure*} 
  \centering
  \includegraphics[width=0.96\textwidth]{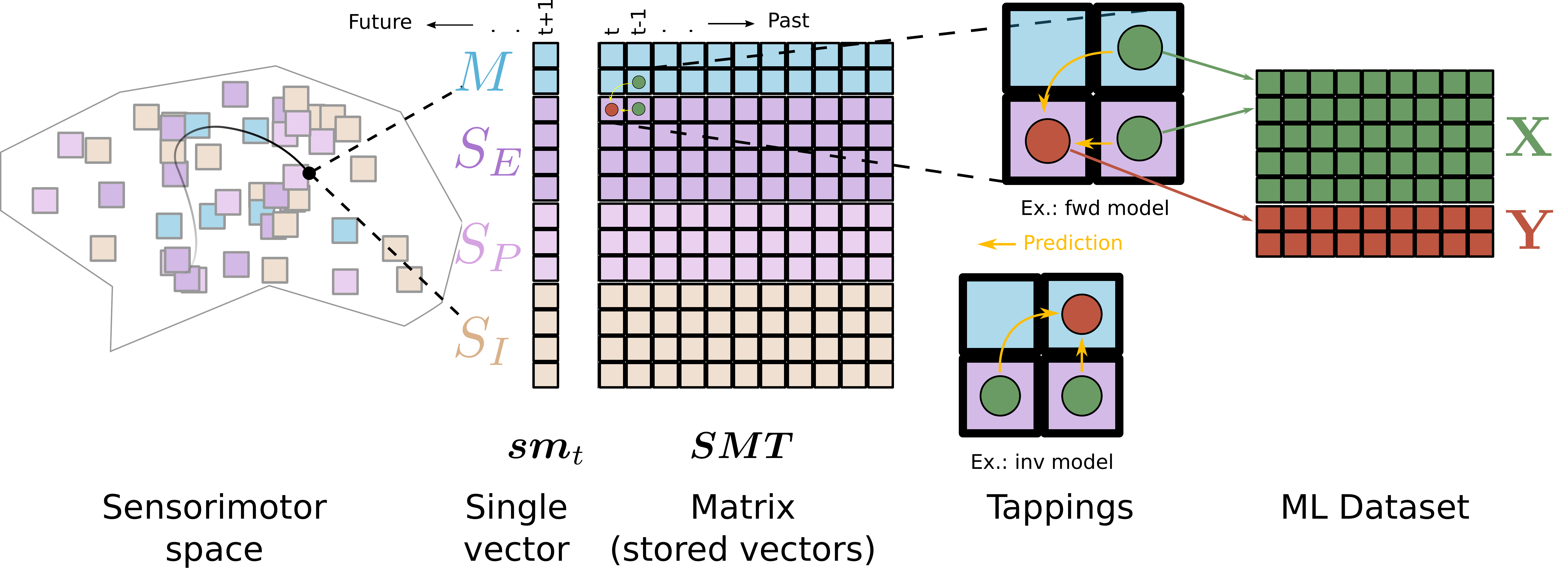}
  \caption{\label{fig:basic-idea} The basic idea of tapping the
    sensorimotor trajectory. Concatenating the row vectors horizontally
      creates a matrix. The matrix inherits the row structure from the
      vector and represents time along the other axis.}
\end{figure*}

The paper is structured as follows. In section \ref{sec:org8f62a06}, related work and existing concepts are discussed.
Section \ref{sec:model_basics} explains the basics of tappings, and in section \ref{sec:org1df4029} specific examples are 
discussed. In section \ref{sec:tap-common} some particular application areas for tappings are named. The paper ends with a discussion and
conclusions.

\section{Related work}
\label{sec:org8f62a06}





\myparb{Signal processing} A central concept in signal processing are
linear filters. These were originally implemented as analog circuits
using \emph{delay lines} to store a finite amount of the signal's past
values. In time-discrete implementations a filter's output is computed
as a weighted sum over a finite number of past inputs. This is
realized by \emph{tapping} into fixed positions within a sliding
window. Each tap is multiplied by a corresponding weight which
together comprise the filter's coefficients. This provides the
starting point for sensorimotor tappings. A filter can be seen as
linear regression and its coefficients can be learned with a least
squares fit. This is known as an adaptive filter in signal processing
and is the same as a linear adaptive forward model in a developmental
robotics context.

\myparb{Modelling tools} The main techniques used for describing
developmental models are plain text accounts, equations, and various
types of block diagrams. Equations and diagrams are each highlighting
different aspects of a model's function and behaviour. Equations are
precise in representing functional dependencies including
general temporal relations. Block diagrams emphasize which functions are
used and which of those functions are interacting directly. None of
them provides an intuitive representation of the global extent and the
microstructure of interaction between variables for a given robot. This
also means that reoccurring patterns of these properties and their
systematic variation across different robots are hard to express.







More systematic graphical methods are the backup diagrams introduced
by Sutton \& Barto \cite{Sutton98} and temporal probabilistic
graphical models \cite{Koller:2009:PGM:1795555}. Backup diagrams track
how the instantaneous information is related to previous states and
indicates how it is propagated back in time to update the relevant
state in the agent's controller. These diagrams do not however
differentiate sensory modalities very well. Probabilistic graphical
models, especially dynamic bayesian networks, provide a natural
complement to the current approach. Like recurrent neural networks,
these models incorporate the problem of mapping input time and
modality into the model state. In contrast, tappings aim at a
decoupled representation of the input mapping and the model's state
update.












\myparb{Information theory} Information theory can be used to quantify
the amount of \emph{shared} information among sensorimotor variables
as shown in \cite{Lungarella2005} or
\cite{doi:10.1163/156855306778522514}. This provides the empirical
complement of tappings and can be used to obtain a tapping from data
\emph{prior} to training a model or to analyze a model's use of
temporal information after training. A number of recent works have
suggested \emph{predictive information}, the amount of information
shared between the past and the future of a random variable, as a
measure for the amount of non-trivial information obtained from
embodied interaction \cite{1999cond.mat..2341B}. This also highlights
the importance of the agent's momentary temporal sensorimotor
embedding.









\myparb{Contribution} Internal modelling approaches in
developmental robotics that use prediction learning are lacking a way
to describe the interaction of the
embedded sensorimotor models with the information provided by the enclosing
developmental model in a general and systematic manner. 
This also holds for temporal difference learning
in RL and correlational learning processes in neuroscience. Thus we
see a definite need for an additional tool from which these fields,
and maybe robotics and AI at large, might benefit. Our contribution 
besides the identification of this gap is a proposal for filling it.






\section{Tappings}
\label{sec:model_basics}






\myparb{Sensorimotor space} The sequence of steps necessary for going
from sensorimotor space to the sensorimotor model input / output space
are shown in the illustration in \autoref{fig:basic-idea} with
enlarged views of two example tappings. A single sensory measurement
at time $t$ is represented by a vector. The vector is composed of
subparts that reflect the natural structure of the agent's
\emph{modalities} imposed by the sensors (e.g. vision or joint
angles). The set of all possible vectors defines the agent's
sensorimotor space. Measurement vector and sensorimotor space comprise
the left part of the figure. The agent's internal time creates the
temporal ordering of incoming measurements
\cite{10.3389/frobt.2016.00004}, and storing them in this order forms
a matrix. The matrix is shown in the center of the figure. It contains
a numerical representation of the sequence of external states \emph{as
  they are reflected} in sensorimotor space. An agent living in a
partially observable world can benefit from extracting additional
information from relations across time and modalities. To do this with
memoryless models, the sensorimotor matrix has to be tapped using a
context dependent pattern attached to the current time step with the
data sliding along underneath. The patterns for a forward and an
inverse model are shown close up. The locations of the nodes of the
tapping indicate which relative time step and modalities are used to
assemble a supervised training set. The node's colors indicate wether
the datum is an input or a target.

\subsection{Example}

\begin{figure}
  \centering
  \includegraphics[width=0.4\textwidth]{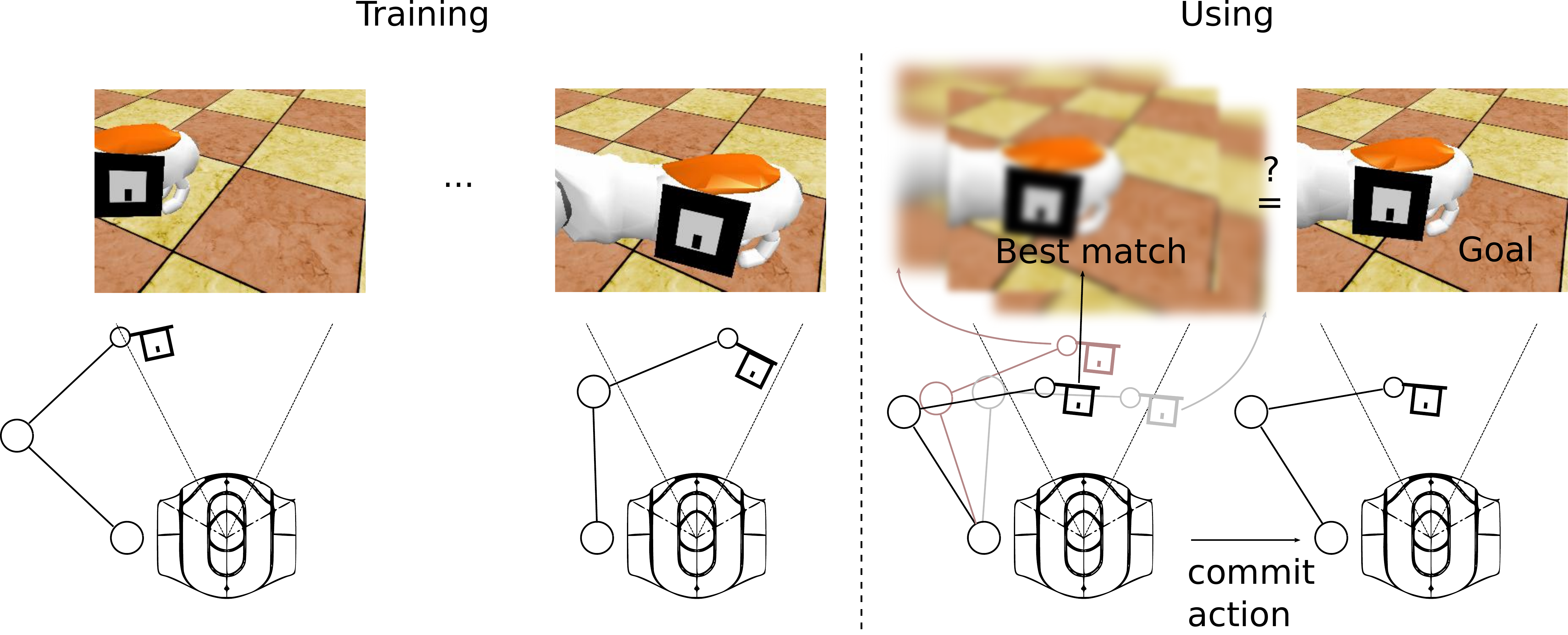}
    \caption{On the left a Nao robot trains a model to predict visual
      consequences from joint angle configurations through
      sensorimotor exploration, right: the robot uses the model to
      find the best matching prediction and the associated action in
      the predictor's input.}
    \label{fig:ex-1-nao-hand-three}
\end{figure}

\begin{figure}
  \centering
  \includegraphics[width=0.4\textwidth]{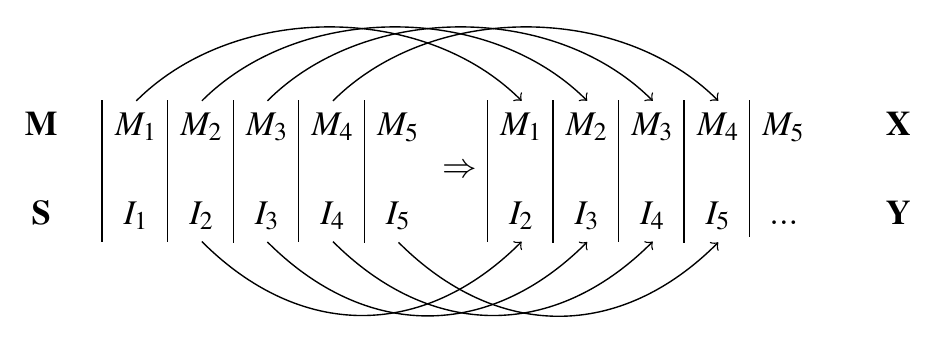}
    \caption{An unrolled view of the repeated application of a tapping into
      sensorimotor data that the Nao agent uses for constructing the
      training data with inputs $\bm{X}$ and targets $\bm{Y}$.}
    \label{fig:ex-2-nao-remapping}
\end{figure}

\begin{figure}
  \centering
  \begin{subfigure}[b]{0.48\textwidth}
    \includegraphics[width=\textwidth]{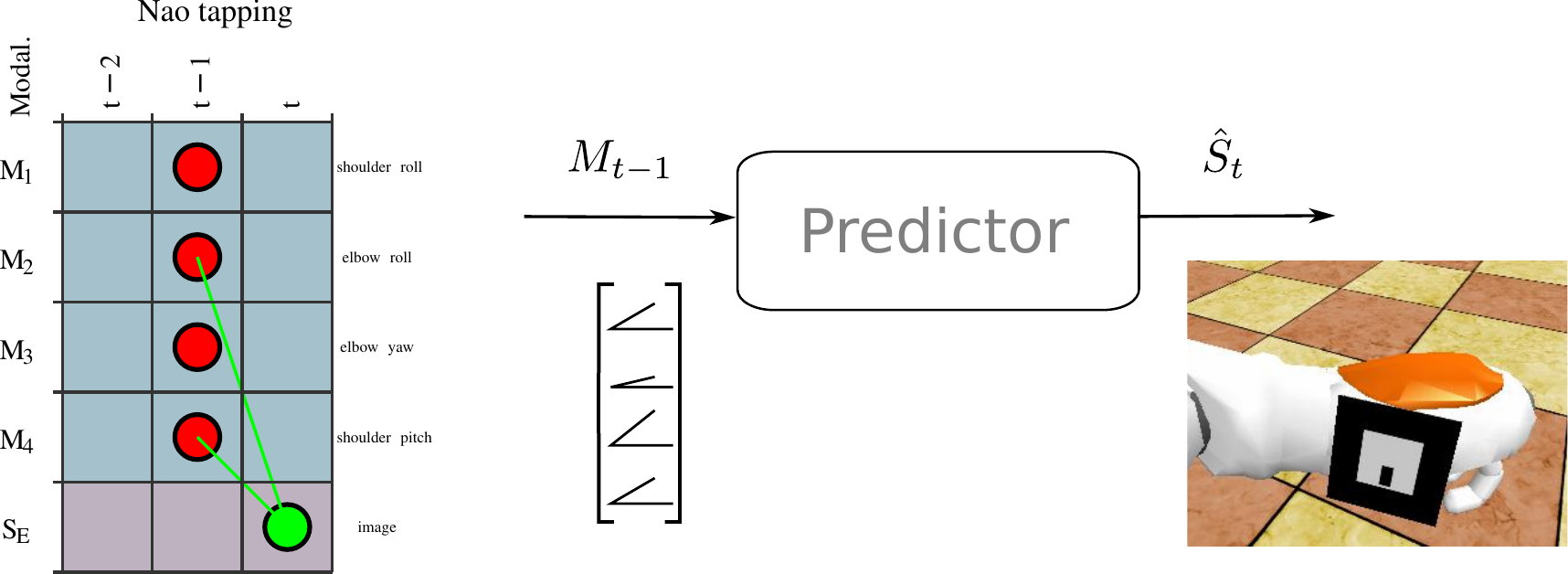}
  \end{subfigure}
  \caption{Tapping for the Nao example with fully expanded motor
    signals and a corresponding block diagram.}
  \label{fig:ex-3-nao-tapping}
\end{figure}

Consider the example of a Nao robot bootstrapping the ability to move
its hand to a given point in visual space shown in
\autoref{fig:ex-1-nao-hand-three}. The agent creates an episode of
data by exploring five random joint angles. For simplicity a kinematic
arm is assumed so there is a delay of one time step between motor
command and the corresponding measurement. Each
momentary measurement consists of the
\emph{current} image, resulting from the previous command, and a
\emph{new} motor command about to be committed. In order to let the
agent learn to predict the image in the next time step from the
current command, an adaptive model is trained with commands as input
and the image as target taken from different relative time steps as shown in
\autoref{fig:ex-2-nao-remapping}. The training set is created from the
raw data by shifting the row of commands one time step to the
right. The measurements in each column of the new matrix are now
ordered by model update steps instead of sensorimotor time. A
detailed tapping is shown \autoref{fig:ex-3-nao-tapping}.



\subsection{Tapping degrees of freedom}
\label{sec:org0564aaf}

\myparb{Column structure} Tappings are specified relative to the
current time $t = 0$, becoming positive in the future and negative
into the past. This proposal only considers discrete time and
equidistant sampling with a constant \(\Delta t\). \myparb{Row
  structure} It makes sense to group variables in the matrix according
to their modality such as as exteroceptive- (vision, hearing),
proprioceptive- (motors, joint angles, forces), or
\emph{interoceptive} sensors. Interoceptive variables represent any
intermediate stage of other concurrent computations in the agent's
sensorimotor loop. A group whose elements all contribute to the same
argument of the target function, for example all pixels in an image,
can be reduced to a single element in the graphical representation.



A common arrangement in a developmental model is to use a supervised
learning algorithm because it can be trained effectively. A supervised
training set consists of the input $\bm{X}$ and targets $\bm{Y}$ that
constrain the functional relation $f(\bm{X}) = \bm{Y}$. The
approximation task is to find parameters $\theta$ for the model $\hat
f(\cdot, \theta) = \bm{\hat Y}$ such that $|\bm{Y} - \bm{\hat Y}|$ is
minimized under a given loss. Prediction learning allows the agent to
construct infinite supervised training data on the fly. Tappings can
describe the necessary transformations independent of the learning
algorithm. If $\bm{XY}$ is the full supervised training set, the
tapping defines a map taking an $\bm{SMT}$ index set to an $\bm{XY}$
index set.




It can be immediately seen from the figures that a tapping is a
directed graph on top of $\bm{SMT}$'s row and column indices. The
graphical structure encodes the relation prescribed by the
sensorimotor model's function inside the developmental model. In
addition to the supervised learning case the graph can immediately be
taken as dynamic Bayesian network graph connecting the current
approach to a rich existing body of formalism and inference
techniques.

\begin{figure}
  \centering
  \begin{subfigure}[b]{0.45\textwidth}
    \includegraphics[width=\textwidth]{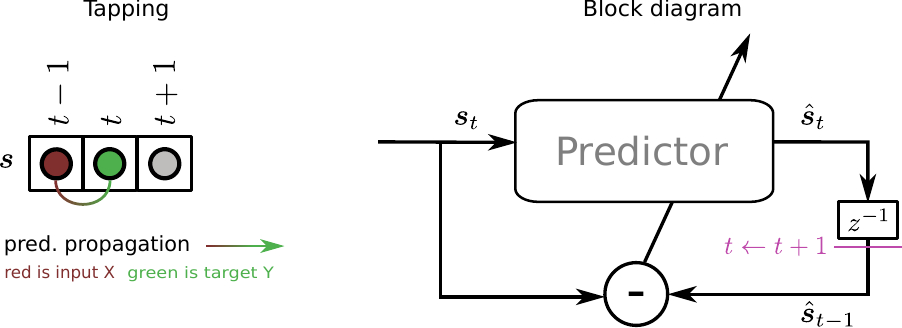}
    \caption{Simple temporal predictor.}
    \label{fig:ex-tpg-simplepred}
  \end{subfigure}
  \vspace{2pt}
  \begin{subfigure}[b]{0.45\textwidth}
    \includegraphics[width=\textwidth]{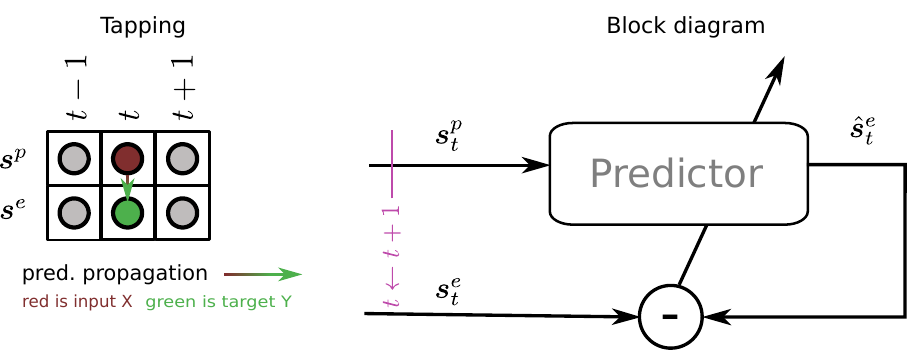}
    \caption{Simple intermodal predictor.}
    \label{fig:ex-tpg-intermodal}
  \end{subfigure}
  \caption{The two principal axes of association shown as tappings
    alongside with corresponding block diagrams. a) A simple temporal
    predictor, predicting the state one timestep ahead, and b) a
    simple intermodal predictor taking proprioceptive input to an
    exteroceptive prediction.}
\end{figure}


\myparb{Time} 
A single time step prediction problem requires a tapping from one time
step to the next. \myparb{Modality} Doing the same along modalities
captures intermodal prediction, that is, predicting sensory
consequences in one modality from the state of another modality. By
adding joint angle sensors to the Nao agent, it could learn to predict
the hand position (vision) from joint angles (proprioception) in the
same time step. 







The two dimensions of the sensorimotor data matrix result in two
corresponding tappings resulting in a temporal predictor, shown in
\autoref{fig:ex-tpg-simplepred}, and an intermodal predictor shown in
\autoref{fig:ex-tpg-intermodal}. Sensorimotor models encode regularity
in sensorimotor state transitions along these axes. \myparb{Forward
  change} Learning transitions along the normal forward flow of time
results in a forward model. Forward models are central to the
simulation theory of cognition, which states that an agent 
learning to approximate the forward transition rules to a sufficient
degree of fidelity can use them to internally ``simulate experience''
\cite{Hesslow201271}.









\myparb{Backward change} Rearranging the direction of prediction to go
backwards in time creates an inverse model. This allows the model to
predict (infer) causes from observed effects, which allows the agent
to control and change its own state by directly predicting the
causes of its desired state. This translates to predicting the actions
that lead to a goal \cite{Rolf:2015}. Direct prediction imposes 
constraints on the learning algorithm. Generally the inverse of a
function can be a correspondence, requiring the learning algorithm to
be able to represent this type relation.

\subsection{Summary}
\myparb{Tappings summary}To summarize this section we highlight the
main features of tappings. They provide an information centric view on
developmental models. This view is independent of particular learning
algorithms, and it provides an upper bound\footnote{the joint entropy
  of all sensorimotor variables} on the amount of explanation a model
needs to accomplish. That bound is a reference for comparing different
models in terms of the fraction of maximum explanation. Tappings
facilitate the design of developmental models, algorithms and their
implementations by highlighting regularities in the design space and
being precise and explicit about time. Analysing two important model
types and their tappings shows to what extent different functional
roles are determined by the input / output relations, and the learning
algorithm respectively. These features all contribute to facilitate
systematic exploration of developmental models.

\section{Basic tappings}
\label{sec:org1df4029}



\begin{figure}
  \centering
    \includegraphics[width=0.45\textwidth]{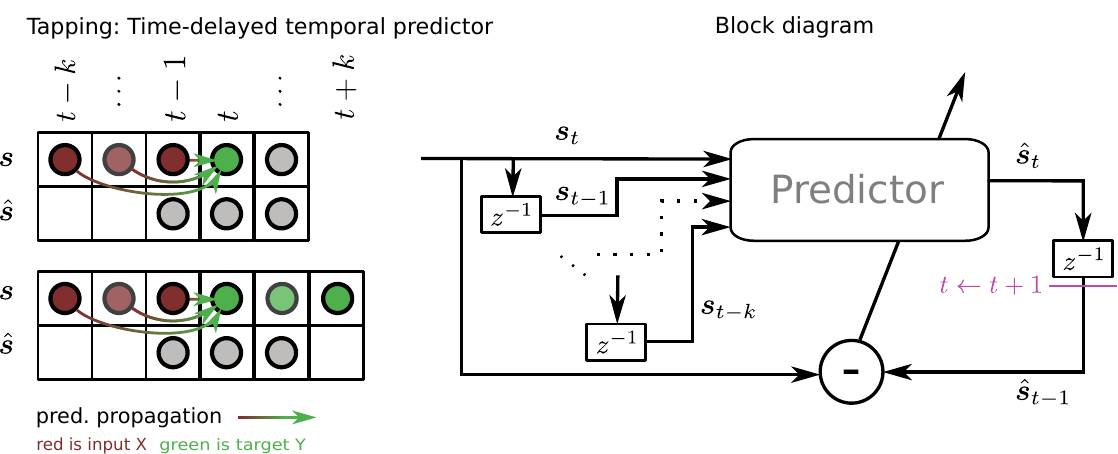}
    \caption{The multi step predictor using a window on $k$ past values as
      instantaneous input and, in the fully symmetric case a window
      on $k-1$ additional future values as the target. The time
      indexing has been omitted for simplicity.}
    \label{fig:ex-tpg-multipred}
\end{figure}

\myparb{Variations} In this section we explore tappings further by
looking at some variations of the simple ones that came out of the
previous section: multi step prediction, autoencoding, and
autopredictive encoding. \myparb{Simple predictor} If the internal
model is a feedforward map without internal memory the simple one time
step predictor in \autoref{fig:ex-tpg-simplepred} cannot make use of
additional information about the future that was presented more than
one time step ago. The missing memory
of the model can be replaced by using a \emph{moving window} of
size $k$ that augments the momentary model input by including all $k$
previous values of the variables\footnote{The moving window technique
  is alternatively known as moving average model, time delay neural network,
delay-embedding or method of delays}. Since tappings are moving
windows, the multi time step tapping shown in
\autoref{fig:ex-tpg-multipred} is almost trivial, the window size
being equal to the number of input taps spread uniformly into the
past. Iterative predictions in extended
forward simulations demand better model accuracy. A reasonable
shortcut towards more accuracy is to improve the prediction by
imposing a long-term consistency constraint by extending the target
tapping into the future (using buffering in closed-loop learning).






\begin{figure}
  \centering
    \includegraphics[width=0.35\textwidth]{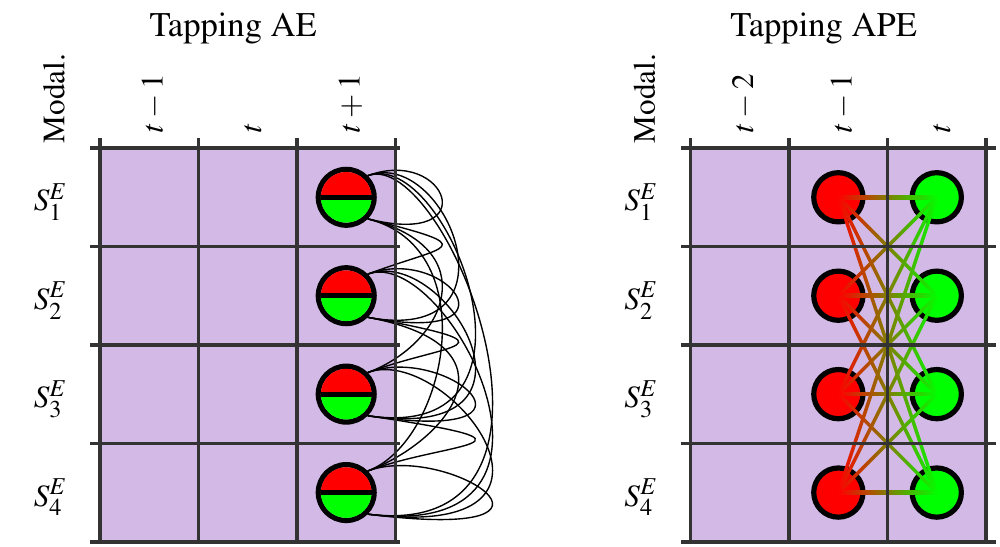}
    \caption{Autoencoder (left) and autopredictive encoder
      (right). The AE's tapping is special because input and target
      coincide. Pulling the input and source apart over one timestep
      difference produces the autopredictive encoder. The prediction
      prior imposes additional structure on the hidden representation.}
    \label{fig:ex-tpg-ae-ape}
\end{figure}

\myparb{Autoencoder} A special case of a predictor is the
autoencoder. Its tapping is shown on the left in
\autoref{fig:ex-tpg-ae-ape}. Its target output is the same as its
input. In terms of the $\bm{XY}$ formulation with $\bm{X} = \bm{Y}$,
the autoencoder could only consist of wires. The added value of an
autoencoder comes exclusively from constraints on the intermediate
representation. Like prediction learning, autoencoding is an
unsupervised learning technique built with supervised learning. 
If we look at the tapping we see
that the information of each single variable on the input is
distributed to all other variables on the
output. \myparb{Auto-Predictive-Encoder} By a simple change of the
tapping we easily obtain an autopredictive encoder (APE) as the result
of pulling the autoencoder's input and output taps one time step
apart. The autopredictive encoder is not an established term but
multiple proposals for such architectures have in fact been made
\cite{NIPS2014_5549,DBLP:journals/corr/PatrauceanHC15,copete16:_motor}.
Applying the prediction constraint on the model has been shown to
increase the task-independence of latent space representations in
\cite{DBLP:journals/corr/LotterKC15}. In
the tapping we see immediately that the prediction constraint 
encourages the model to represent the rules of change in the hidden
space. The APE tapping is shown in
\autoref{fig:ex-tpg-ae-ape}.

\section{Application areas}
\label{sec:tap-common}



\begin{figure}
  \begin{subfigure}[b]{0.45\textwidth}
    \includegraphics[width=\textwidth]{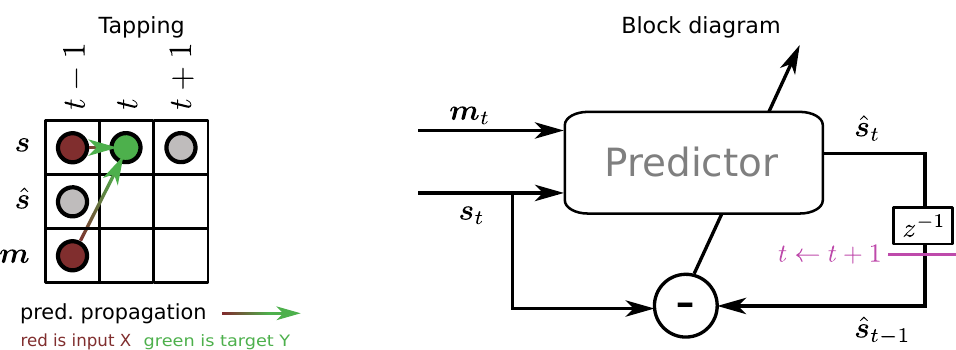}
    \caption{Forward}
    \label{fig:ex-tpg-mdl-fwd}
  \end{subfigure}
  \quad
  \begin{subfigure}[b]{0.45\textwidth}
    \includegraphics[width=\textwidth]{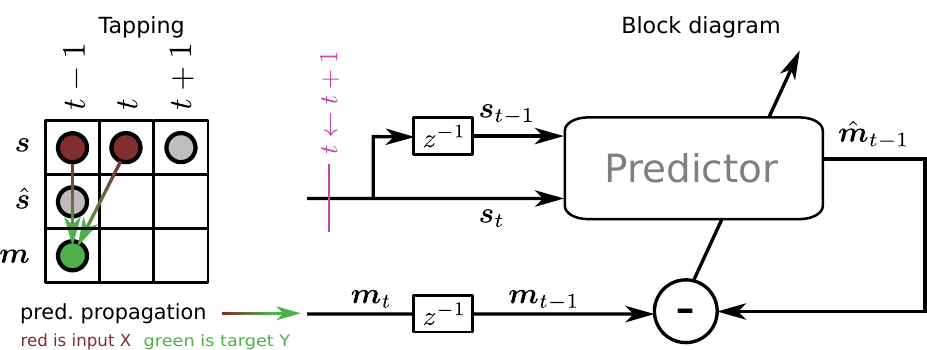}
    \caption{Inverse}
    \label{fig:ex-tpg-mdl-inv}
  \end{subfigure}
  \caption{Tapping a single time step forward- and inverse model
    pair. The model's functions are determined by different relations
    over the same set of variables.}
    \label{fig:ex-tpg-mdl-pair}
\end{figure}

\myparb{Developmental robotics} Internal modelling
\cite{craik43:_natur_explan} is an important concept used in
developmental robotics
\cite{Wolpert98,Demiris2006361,10.3389/frobt.2016.00039}.  An
underlying driving hypothesis is that predictive models enable
\emph{anticipatory} behaviour \cite{rosen12:anticipatory_systems}
which is more powerful than purely reactive behaviour. From the
developmental perspective this implies that some functions of a
developmental model must be provided by adaptive models of the
sensorimotor dynamics. Two basic functional types of internal models,
forward and inverse ones, have already been introduced as examples in
\autoref{fig:basic-idea} and are shown again as a pair of tappings in
\autoref{fig:ex-tpg-mdl-pair}. This highlights the rearrangement of
the direction of prediction without a change of
variables. Exploitation of adaptive models has also been described
above indicating different ways of predicting and evaluating future
options with forward models, or directly inferring actions with inverse
models.


\myparb{Reinforcement learning} A popular method in reinforcement
learning is temporal difference learning. Temporal difference learning
is a family of algorithms to approximate a prediction target with a
recurrent estimate. The usual target is a \emph{value function} which maps
actions to a value. The estimate is bootstrapped by minimizing the
moment-to-moment value prediction error, which is ultimately grounded
in a primary reward signal. There exists
extensive theory in RL that deals with the problem of integrating
task-relevant information that is spread out in time, with two
fundamental concepts being involved. The first one
is that of \emph{multistep methods} which take care of consequences escaping
into the future. The second one are \emph{eligibility traces} which
capture causes vanishing into the past. Taken together they solve the
general delayed reward problem. Depending on the parameters a
corresponding tapping will be similar to the multi step predictor.

\begin{figure}
    \centering
    \includegraphics[width=0.45\textwidth]{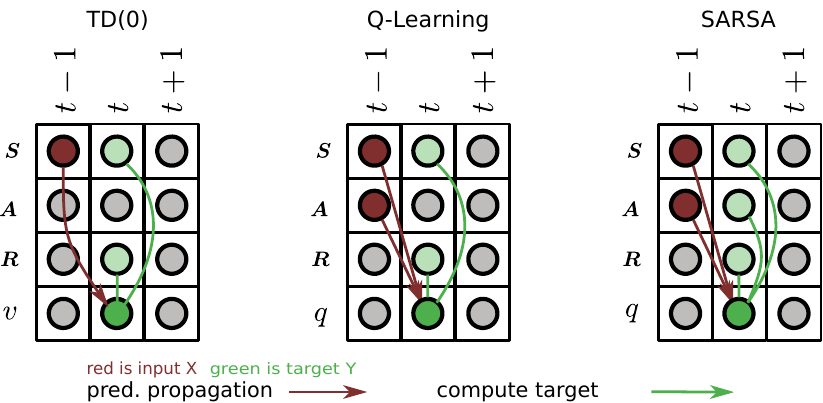}
    \caption{Tapping temporal difference learning algorithms.}
    \label{fig:ex-tpg-rl-td}
\end{figure}

The importance of features and modalities and the information
contained in their mutual relations is less developed. The concepts
used in reinforcement learning can easily be remapped to internal
modelling terms and vice versa, making tappings immediately applicable
to temporal difference learning problems. Looking at three basic
temporal difference learning algorithms, TD(0), Q-Learning and SARSA,
it can be seen that they all approximate a target by updating from a one time
step difference. TD(0)'s target is a state value function $v$ while
for Q-learning and SARSA it is a state-action value function $q$
\cite{Sutton98}. The update rules all follow the same general form of

\[
\Delta v = \alpha (R_t + \gamma v(S')_t)
\]

and the corresponding tappings are shown in
\autoref{fig:ex-tpg-rl-td}. Comparing these with the internal model
tappings we see that temporal difference learning corresponds with
prediction learning and that the value function is a forward model
allowing us to reframe RL problems as developmental prediction
learning ones and the other way round. The $\lambda = 0$
case is shown here to correspond to a single time step
tapping but the proportional increase in tapping length with
increasing $\lambda$ should be obvious.








\begin{figure}
    \centering
    \includegraphics[width=0.19\textwidth]{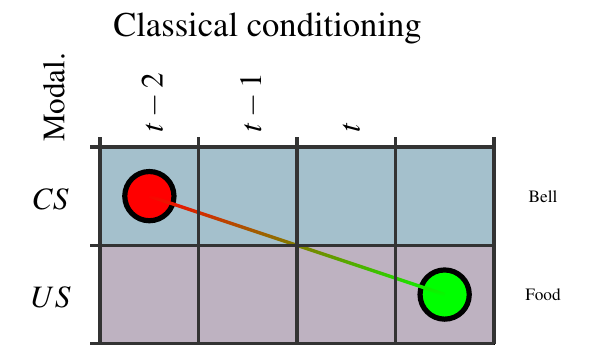}
    \caption{Model of classical conditioning: it explains the
      prediction of the unconditioned stimulus (US) from a stimulus
      occurring earlier in time, the conditioned stimulus (CS). The
      predictive association of stimuli across time is precisely the
      process of conditioning. This highlights again that the
      difference to a forward model or a value prediction is only in
      the terminology and not in the structure of the association
      problem.}
    \label{fig:ex-tpg-conditioning}
\end{figure}

\myparb{Neuroscience} Neuroscience provides several models that link
computational and neurobiological accounts of associative learning and
reinforcement learning. The Rescorla-Wagner rule
\cite{rescorla72:_pavlov} is one example. It is a model of classical
conditioning and describes how an association is learned across two
modalities, the unconditioned (US) and the conditioned stimulus (CS),
which occur at different times. Another example is the reward
prediction error hypothesis of dopamine
\cite{Schultz1593,Dayan02,Niv2009139} which provides a physiological
mechanism in support of computational descriptions of
reinforcement. Low-level models of neural adaptation like spike-time
dependent plasticity (STDP) \cite{gerstner96,Markram213} are
characterized by a local window of interaction on a microscopic time
scale. STDP itself is not a model for learning delays but an even
lower level mechanism for reinforcing or weakening the association of
pre- and post-synaptic events based on the local window prior. It can
of course be used indirectly to extract sensorimotor delay
information. Tappings apply without modification to all these
different levels of modeling as shown exemplarily for the conditioning
case in \autoref{fig:ex-tpg-conditioning}.



\section{Discussion}
\label{sec:discussion}

During this presentation of tappings, a few additional issues came up that still need to be discussed.
\myparb{Recurrence and autoregression} Models with memory like
recurrent neural networks or dynamic Bayesian networks need special
consideration with respect to tappings. Such models naturally retain
an internal memory of past input values. Because of this, they do not
need explicit memory in their inputs and in theory only need to tap
across one time step. They are building up an implicit tapping as part
of their learning while tappings aim at an representation of specific
memory needs for a given learning task. Measuring the information flow
across the model inputs and outputs after training with quantitative
\cite{Lungarella2005} or relational techniques
\cite{2010arXiv1004.2515W} should result in an effective tapping that
could be used for comparison with prior tappings or interpreted as a
way of learning them.





\myparb{Info loss} The memory issue is an example of a more general
aspect about tappings. The current proposal disregards details about
the learning algorithm used at the level of sensorimotor models. It is
argued that this is in fact an advantage and necessary for wider
comparison of models. The same is evident in the case of inverse
problems where the learning of correspondences instead of functions
needs to be considered. It remains to be shown how these properties
could be integrated and represented in a tapping.

No experimental validation is given in this paper. This does not mean
though that no experimental backing of the idea exists. In fact, it is
an outcome of analyzing the design of a large number of different
experiments in developmental learning in the course of our research
during which the structural invariant that tappings try to represent
became evident. It was decided to defer the experimental validation in
this paper in favour of being able to focus on the explanation of the
basic idea and to demonstrate its applicability to a wide range of
contexts.

\section{Conclusion}
\label{sec:org91c9ccc}



\myparb{Summary} This paper introduces tappings, a novel concept for
designing and analysing models of developmental learning in the field
of developmental robotics and the related fields of reinforcement
learning and computational neuroscience. Tappings came out of a need
for capturing the detailed embedding of learning machines in the
temporal and modal context of raw sensorimotor trajectories. Tappings
create a particular view on the interaction between the embodiment and
the functional requirements of behaviour that can help to better
understand developmental learning processes, and make sensorimotor
learning more efficient. They can systematically describe the
relationship between supervised learning and developmental models. By
ignoring computational details the tapping view highlights the
information flow across models and using that we can compare a large
range of models that cannot easily be compared otherwise. We showed
the structural similarity of prediction learning in the developmental
context and temporal difference learning in RL.

\section*{Acknowledgment}

We would like to thank Guido Schillaci for providing experimental data
for the Nao example and the Adaptive Systems Group members for discussions.

\IEEEtriggeratref{20}
\IEEEtriggercmd{\enlargethispage{-5in}}



\bibliography{tappings}{} 
\bibliographystyle{IEEEtran}

\vspace{10pt}
\vspace{10pt}
\vspace{10pt}
\vspace{10pt}
\vspace{10pt}
\vspace{10pt}

%





\end{document}